\documentclass[runningheads,a4paper]{llncs}
\usepackage[pass]{geometry}

\usepackage{graphicx}
\usepackage{epstopdf}
\usepackage[latin1]{inputenc}
\usepackage{amsfonts}
\usepackage{amssymb}
\usepackage{graphicx}
\usepackage{listings}
\usepackage{url}
\usepackage{algorithm2e}
\usepackage{gensymb}
\usepackage{textcomp}

\usepackage[overlay,absolute]{textpos}

\begin{document}

\begin{textblock*}{10in}(38mm, 10mm)
{\textbf{Ref:} \emph{International Conference on Artificial Neural Networks (ICANN)}, Springer LNCS,}
\end{textblock*}
\begin{textblock*}{10in}(38mm, 15mm)
{Vol.~9887, pp.~20--28, Barcelona, Spain, September 2016.}
\end{textblock*}

\author{Eli (Omid) David\inst{1} \and Nathan S. Netanyahu\inst{1,2} }

\authorrunning{E.O. David, N.S. Netanyahu}

\institute{
Department of Computer Science, Bar-Ilan University, Ramat-Gan, Israel \\
\email{mail@elidavid.com, nathan@cs.biu.ac.il}
\and
Center for Automation Research, University of Maryland, College Park, MD, USA\\
\email{nathan@cfar.umd.edu}
}

\title{DeepPainter: Painter Classification Using Deep Convolutional Autoencoders}

\titlerunning{DeepPainter: Painter Classification Using Deep Convolutional Autoencoders}

\maketitle
\vspace{-10pt}

\begin{abstract}
In this paper we describe the problem of painter classification, and propose a novel approach based on deep convolutional autoencoder neural networks. While previous approaches relied on image processing and manual feature extraction from paintings, our approach operates on the raw pixel level, without any preprocessing or manual feature extraction. We first train a deep convolutional autoencoder on a dataset of paintings, and subsequently use it to initialize a supervised convolutional neural network for the classification phase.

The proposed approach substantially outperforms previous methods, improving the previous state-of-the-art for the 3-painter classification problem from 90.44\% accuracy (previous state-of-the-art) to 96.52\% accuracy, i.e., a 63\% reduction in error rate.
\end{abstract}

\section{Introduction}
\vspace{-6pt}

Art forgery, which dates back more than two thousand years, has played a key role in the development of painting authentication. This task has been usually performed manually by art experts who have dedicated their lives to this profession. Their expertise amounted to using various characteristics other than what the human eye can see, including chemical analysis, spectrometry, and infrared or X-ray imaging. The infamous Vermeer forgery \cite{philipsvermeer} attests, perhaps, most vividly to the challenges presented by painting authentication. Han van Meegeren used historical canvasses and managed to deceive art experts into believing that his painting was an authentic Vermeer. Only after being charged with treason and sentenced to death for selling another (forged) Vermeer, did he confess and was forced to create another painting to prove himself innocent of treason. A more recent case of painting authenticity involves the Pollock paintings found a decade ago in a storage locker in Wainscott, NY. The authenticity of these paintings was compromised on the basis of computer analysis of the paintings' fractal dimension \cite{taylor2007authenticating}. This claim was subsequently disputed by analyzing childlike drawings that supposedly have the same fractal dimension as the Pollock paintings \cite{jones2006fractal}.

In this paper we address the closely related problem of painting classification, i.e., the task of assigning a specific artist to a given painting (from a dataset of paintings by several artists). Note that the image authentication problem can be viewed as a binary image classification problem (i.e., determine whether or not a given painting was painted by a certain artist). Recent developments for both problem types have focused on preprocessing techniques of reducing the high dimensionality of visual data to low-dimensional representations which can be manipulated towards image understanding. 

Levy \textit{et al.} \cite{levy2013painter,levy2014painter} applied feature extraction to paintings using generic image processing (IP) functions (e.g., fractal dimension, Fourier spectra coefficients, texture coefficients, etc.), and restricted Boltzmann machines (RBM), followed by genetic algorithms (GA)-based learning of the weights of a weighted nearest neighbor (WNN) classifier \cite{siedlecki1989note}. Their approach achieved 90.44\% classification accuracy for the 3-painter classification problem.

In this paper we present the problem of painter classification and briefly survey recent research that has been conducted in the field. We then present our novel approach, which uses convolutional autoencoders (CAE) instead of image processing based feature extraction. We subsequently use the trained CAE to initialize a convolutional neural network (CNN) for supervised training on specific painters. The results demonstrate a substantial improvement over previous methods, improving the accuracy to 96.52\%. This sets a new state-of-the-art for the painter classification problem,

\section{Background}
\vspace{-6pt}

Image authentication is the task of determining whether or not
a given painting was painted by a specific artist. The related
task addressed by us, though, is image classification,  i.e., the task of determining the artist of a given painting (from a certain group of artists). The input to our problem
consists of painting images of the group of artists (several paintings of each artist), and our objective is to automatically
classify a given painting. One of the difficulties in solving this problem is that we cannot define a certain set of rules that the painting has to conform to in order to classify it to the subgroup corresponding to the correct artist. For this reason, computer vision techniques which are capable of identifying shapes and objects in an image are not sufficiently effective for solving the problem.

Formerly there have been attempts to harness the strength of image analysis tools to classify historical art paintings into categories of artists or genres. Levy \textit{et al.} \cite{levy2013painter} used GA-based WNN with a set of 78 prevalent image features for classifying paintings by Rembrandt, Renoir, and van Gogh, obtaining 80\% classification accuracy. In their later work \cite{levy2014painter}, they augmented their approach by also adding 20 features using restricted Boltzmann machines (RBM)\cite{hinton2006fast}, improving the classification accuracy to 90.44\%. 

Herik and Postma \cite{van2000discovering} surveyed
image features relevant to the historic art domain and concluded that neural network techniques combined with domain knowledge were most suitable to the task of automatic image classification. Under-drawing strokes in infrared reflectograms were analyzed by Kammerer \textit{et al.} \cite{kammerer2007identification} in order to classify how and by what tools paintings are painted. Natural language processing techniques using a naive-Bayes classifier and the coefficients of a discrete cosine transform (DCT) were used by Keren \cite{keren2002painter} in order to classify local features in an image. Kroner \textit{et al.} \cite{kroner1998authentication} classified drawings by using image histograms and pattern recognition methods.

The above past research focused on specific image processing features tailored for specific datasets (such as ink paintings, infrared reflectograms, or black and white sketches). This domain-specific knowledge facilitates the exploitation of various characteristics of the painting-specific domain. 

In the next section we present our convolutional autoencoder based approach, which does not incorporate any domain-specific knowledge, and in fact is operating solely on the raw pixel level.

\section{Feature Extraction Using Convolutional Autoencoders}
\label{features_extracted}
\vspace{-6pt}

\subsection{Convolutional Neural Networks}

\begin{figure}[ht]
	\centering
	\includegraphics[width=\textwidth]{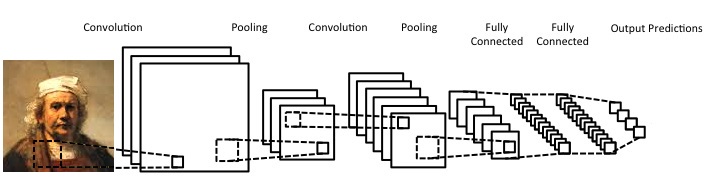}
	\caption{Typical architecture of a convolutional neural network.}
	\label{fig:cnn}
\end{figure}

In recent years convolutional neural networks (CNN)\cite{behnke2003,lecun1988,krizhevsky12} have outperformed conventional image processing methods in all computer vision related tasks they have been applied to. The architecture of a CNN typically includes several components which are stacked on top of each other: the convolutional layer, the max-pooling layer, which subsamples the data (e.g., for each $2 \times 2$ region selects only the maximum value, thus resulting in four times reduction in size), and finally a classification layer (and usually several fully connected layers before the classification layer). Figure~\ref{fig:cnn} shows a typical CNN,

Standards CNNs are usually used in a supervised framework, where a large training dataset (typically including at least many thousands of images per class) is available. Thus, using CNNs for end-to-end painter classification is problematic, due to a smaller number of training samples available per painter (usually from a few tens of paintings to at most a few hundred paintings for more prolific painters).

\subsection{Autoencoders}
\vspace{-3pt}
Where small number of training samples are available, unsupervised pretraining has proven highly effective \cite{hinton2006fast,vincent2008}. Unsupervised training methods using neural networks involve either the use of restricted Boltzmann machines (RBM)\cite{hinton2006fast} which are trained using contrastive divergence, or autoencoders\cite{vincent2008} which are training using standard backpropagation. 

The basic principle for all methods involves receiving an input $\textbf{x}$ and mapping it to a latent representation $\textbf{h}$, using a function $\textbf{h} = \sigma(Wx + b)$, where $\sigma$ is a nonlinear activation function, $W$ is a matrix of weights between the two layers, and $b$ is bias. The autoencoder then tries to reconstruct the original input by $\textbf{y} = \sigma(W'h + b')$. Thus, each training sample $x_i$ is first mapped to a hidden layer $h_i$ and then reconstructed to $y_i$. The autoencoder is trained using backpropagation to reduce this reconstruction error.

\subsection{Convolutional Autoencoders}
\vspace{-3pt}
The principles behind convolutional neural networks and autoencoders can be combined to produce convolutional autoencoders (CAE). Several approaches involving the combination of these methods have been explored in the past, and here we use a CAE architecture along the lines presented in \cite{masci2011,zeiler2011,zeiler2014}.

In order to use CNNs as autoencoders, for each convolutional layer, a corresponding deconvolutional layer should be constructed. Additionally, max-pooling layers result in loss of information, and so an unpooling layer should try to approximately restore the original values. Note that the subsampling due to max-pooling in fact operates as a strong regularizer.

Deconvolution layers can either be equal but transposed to the original convolution layers, or learned from scratch. Often both approaches work equally well in practice. This is similar to standard autoencoders where the weights of the decoder layer $W'$ can either be learned from scratch, or set to the transpose of the encoder layer ($W' = W^T$), this is referred to as \emph{tied weights}.

\begin{figure}[ht]
	\centering
	\includegraphics[width=0.6\textwidth]{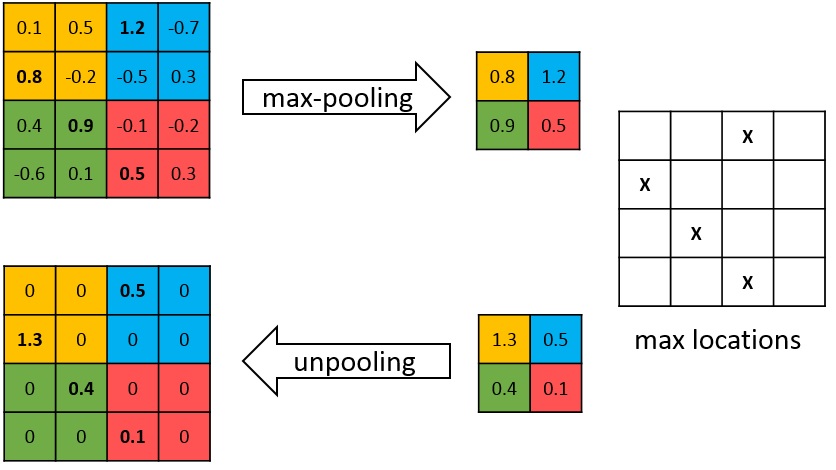}
	\caption{Pooling and unpooling layers. For each pooling layer, the max locations are stored. These locations are then used in unpooling layer.}
	\label{fig:unpooling}
\end{figure}

Several methods have been applied in the past for the unpooling operation \cite{masci2011,zeiler2011,zeiler2014}. Here we employ the method used in \cite{zeiler2011}, where during pooling, the location of maximum value is stored, such that during unpooling the value is restored in that location, and the other locations are set to zero. Figure~\ref{fig:unpooling} illustrates unpooling, and Figure~\ref{fig:cae} shows a complete convolutional autoencoder structure.

After training a CAE, we can remove the unpooling and deconvolution components, and use the convolution and pooling components to initialize a supervised CNN, by adding a fully connected layer followed by a classification layer.

\section{CAE and CNN for Painter Classification}
\vspace{-6pt}

For unsupervised training of CAE, we use a randomly selected set of 5,000 paintings from the Webmuseum (\texttt{webmuseum.meulie.net/wm}). The images have 24-bit color depth with varying resolutions averaged approximately at $1000 \times 1000$ pixels, and compressed as JPEG formatted files. We have resampled the images and normalized them to $ 256 \times 256 $ pixels. The goal here is to train the CAE to find features that are specifically useful for paintings, which have a more specific color and composition range in comparison to real-world images.

\begin{figure}[ht]
	\centering
	\includegraphics[width=\textwidth]{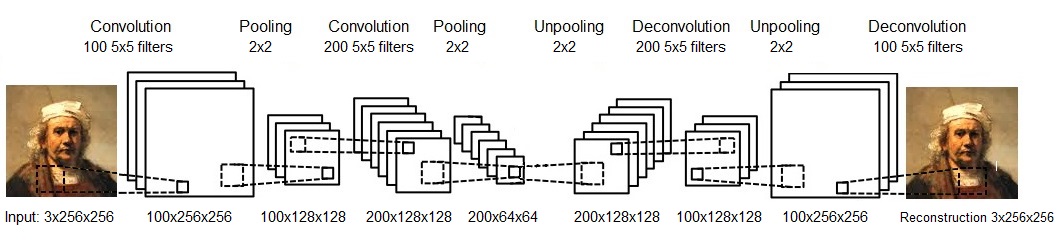}
	\caption{Illustration of convolutional autoencoder. In this example the CAE comprises two convolution layers and their two corresponding deconvolution layers, and two max-pooling layers and their corresponding unpooling layers.}
	\label{fig:cae}
\end{figure}

\noindent Our CAE contains the following layers (see Figure~\ref{fig:cae}). The convolution filter sizes are always of size $5 \times 5$.
\vspace*{-0.5em}
\begin{enumerate}
\item the input layer consists of the raw image (resampled to $256 \times 256 $ pixels) in three channels (R, G, and B)
\item convolutional layer of size $100 \times 256 \times 256$
\item max-pooling layer of size $2 \times 2$
\item convolutional layer of size $200 \times 128 \times 128$
\item max-pooling layer of size $2 \times 2$
\item unpooling layer of size $2 \times 2$
\item deconvolutional layer of size $200 \times 128 \times 128$
\item unpooling layer of size $2 \times 2$
\item deconvolutional layer of size $100 \times 256 \times 256$
\end{enumerate}

The learning rate starts from 0.01 and is multiplied by 0.98 after each epoch. In order to further encourage the CAE to find meaningful features, we randomly remove 20\% of the pixels for the images per epoch. The concept here is similar to that of denoising autoencoders \cite{vincent2008} which outperform traditional autoencoders.

The supervised classification benchmark is identical to that used by Levy \textit{et al.} \cite{levy2013painter,levy2014painter} in their experiments. It consists of $(3 \times 40 =)$ $120$ digital reproductions of paintings by Rembrandt, Renoir, and van Gough, downloaded from the Webmuseum. The Appendix contains the painting titles of the images used in our experiments.

Having trained a CAE, we can now remove the decoder components (items 6 to 9 in the above list) and use the CAE for initializing a supervised CNN. On top of these components due to CNN, we add two fully connected layers of size 400 and 200, followed by a softmax output unit of size three (since there are three painters in the benchmark). The cross entropy loss is used.

\medskip
\noindent The full CNN contains the following layers (see Figure~\ref{fig:cnn}): 
\vspace*{-0.5em}
\begin{enumerate}
\item the input layer consists of the raw image (resampled to $256 \times 256 $ pixels) in three channels (R, G, and B)
\item convolutional layer with 100 $5 \times 5$ filters per input channel
\item max-pooling layer of size $2 \times 2$
\item convolutional layer with 200 $5 \times 5$ filters per map
\item max-pooling layer of size $2 \times 2$
\item fully connected layer of size 400
\item fully connected layer of size 200
\item output softmax layer of size 3
\end{enumerate}

To make our results directly comparable to those of Levy \textit{et al.} \cite{levy2013painter,levy2014painter}, we conducted 10-fold cross validation, where in each of 10 runs 90\% of the data is used for training, and 10\% for validation. 

After performing 10 such training and validation runs, the average accuracy obtained for our CNN over the validation set is 96.52\%. This represents a 63\% reduction in error rate in comparison to the previous state-of-the-art on this benchmark, which stood at 90.44\%.

Table \ref{results} provides a summary of the classification accuracies obtained by previous methods and our method.

\begin{table}[h]
\centering
\setlength{\tabcolsep}{1em} 
{\renewcommand{\arraystretch}{1.2}
\begin{tabular}{|c|c||c|}
\hline
Feature extraction method & Supervised learning method & Accuracy \\ 
\hline
\hline
Image Processing & Nearest Neighbor & 65.71\%\\
\hline
Image Processing & SVM & 68.33\%\\
\hline
Image Processing & Genetic Algorithm & 78.33\%\\
\hline
RBM & Nearest Neighbor & 64.41\%\\
\hline
RBM & SVM & 77.50\%\\
\hline
RBM & Genetic Algorithm & 73.92\%\\
\hline
Image Processing + RBM & Nearest Neighbor & 68.71\%\\
\hline
Image Processing + RBM & SVM & 71.66\%\\
\hline
Image Processing + RBM & Genetic Algorithm & 90.44\%\\
\hline
\textbf{Convolutional Autoencoder} & \textbf{CNN} & \textbf{96.52}\%\\
\hline
\end{tabular} 
}
\medskip
\caption{Classification accuracy for several previous methods and our CAE based method. The results are the average over 10-fold cross validation.}
\label{results}
\end{table}

\vspace*{-20pt}
\section{Conclusion}
\vspace*{-6pt}
Automatic painter classification has gained much attention over the past decades, and much progress has been made with regards to both relevant preprocessing techniques and classification algorithms. Still, the problem of painter classification remains a complex task that requires more sophisticated techniques.

The results presented in this paper show that deep learning methods can be effectively employed for painter classification. Specifically, our results show that convolutional autoencoders are capable of extracting meaningful information from paintings, and combined with supervised convolutional networks, we managed to substantially improve the previous state-of-the-art, from 90.44\% accuracy (previous state-of-the-art) to 96.52\% accuracy, i.e., a 63\% reduction in error rate.

\bibliographystyle{plain}
\bibliography{Painter}

\begin{thebibliography}{10}

\bibitem{behnke2003}
S.~Behnke.
\newblock Hierarchical neural networks for image interpretation.
\newblock {\em LNCS}, 2766:1--13, 2003.

\bibitem{hinton2006fast}
G.E Hinton, S.~Osindero, and Y.W. Teh.
\newblock A fast learning algorithm for deep belief nets.
\newblock {\em Neural computation}, 18(7):1527--1554, 2006.

\bibitem{jones2006fractal}
K.~Jones-Smith and H.~Mathur.
\newblock Fractal analysis: revisiting {P}ollock's drip paintings.
\newblock {\em Nature}, 444(7119):E9--E10, 2006.

\bibitem{kammerer2007identification}
P.~Kammerer, M.~Lettner, E.~Zolda, and R.~Sablatnig.
\newblock Identification of drawing tools by classification of textural and
  boundary features of strokes.
\newblock {\em Pattern Recognition Letters}, 28(6):710--718, 2007.

\bibitem{keren2002painter}
D.~Keren.
\newblock Painter identification using local features and naive {B}ayes.
\newblock In {\em Proceedings of the IEEE International Conference on Pattern
  Recognition}, volume~2, pages 474--477, 2002.

\bibitem{krizhevsky12}
Alex Krizhevsky, Ilya Sutskever, and Geoff Hinton.
\newblock Imagenet classification with deep convolutional neural networks.
\newblock In P.~Bartlett, F.C.N. Pereira, C.J.C. Burges, L.~Bottou, and K.Q.
  Weinberger, editors, {\em Advances in Neural Information Processing Systems
  25}, pages 1106--1114, 2012.

\bibitem{kroner1998authentication}
S.~Kroner and A.~Lattner.
\newblock Authentication of free hand drawings by pattern recognition methods.
\newblock In {\em Proceedings of the IEEE 14th International Conference on
  Pattern Recognition}, volume~1, pages 462--464, 1998.

\bibitem{lecun1988}
Y.~LeCun, L.~Bottou, Y.~Bengio, and P.~Haffner.
\newblock Gradient-based learning applied to document recognition.
\newblock {\em Proceedings of the IEEE}, 86(11).

\bibitem{levy2013painter}
E.~Levy, O.E. David, and N.S. Netanyahu.
\newblock Painter classification using genetic algorithms.
\newblock In {\em IEEE Congress on Evolutionary Computation}, pages 3027--3034,
  2013.

\bibitem{levy2014painter}
E.~Levy, O.E. David, and N.S. Netanyahu.
\newblock Genetic algorithms and deep learning for automatic painter
  classification.
\newblock In {\em ACM Genetic and Evolutionary Computation Conference}, pages
  1143--1150, 2014.

\bibitem{masci2011}
Jonathan Masci, Ueli Meier, Dan Ciresan, and Jurgen Schmidhuber.
\newblock Stacked convolutional auto-encoders for hierarchical feature
  extraction.
\newblock In {\em International Conference on Artificial Neural Networks,
  ICANN}, pages 52--59, 2011.

\bibitem{philipsvermeer}
D.~Phillips.
\newblock How do forgers deceive art critics?
\newblock {\em The Artful Eye}.
\newblock R. Gregory, J. Harris, P. Heard, and D. Rose, Eds., Oxford University
  Press, pages 372--388, 1995.

\bibitem{siedlecki1989note}
W.~Siedlecki and J.~Sklansky.
\newblock A note on genetic algorithms for large-scale feature selection.
\newblock {\em Pattern Recognition Letters}, 10(5):335--347, 1989.

\bibitem{taylor2007authenticating}
R.P. Taylor, R.~Guzman, T.P. Martin, G.D.R. Hall, A.P. Micolich, D.~Jonas, B.C.
  Scannell, M.S. Fairbanks, and C.A. Marlow.
\newblock Authenticating {P}ollock paintings using fractal geometry.
\newblock {\em Pattern Recognition Letters}, 28(6):695--702, 2007.

\bibitem{van2000discovering}
H.J. van~den Herik and E.O. Postma.
\newblock Discovering the visual signature of painters.
\newblock {\em Future Directions for Intelligent Systems and Information
  Sciences}.
\newblock N. Kasabov, Ed., Physica-Verlag, Heidelberg, pages 129--147, 2000.

\bibitem{vincent2008}
P.~Vincent, H.~Larochelle, Y.~Bengio, and P.A. Manzagol.
\newblock Extracting and composing robust features with denoising autoencoders.
\newblock In {\em 25th international Conference on Machine learning, ICML},
  pages 1096--1103, 2008.

\bibitem{zeiler2014}
M.D. Zeiler and R.~Fergus.
\newblock Visualizing and understanding convolutional networks.
\newblock In {\em European Conference on Computer Vision, ECCV}, pages
  818--833, 2014.

\bibitem{zeiler2011}
M.D. Zeiler, G.W. Taylor, and R.~Fergus.
\newblock Adaptive deconvolutional networks for mid and high level feature
  learning.
\newblock In {\em International Conference on Computer Vision, ICCV}, pages
  2018--2025, 2011.

\end{thebibliography}

\newpage
\appendix

This appendix lists the $(40 \times 3 )= 120$ titles of the paintings experimented with by van Gogh, Rembrandt, and Renoir.
\newline

\begin{tabular}{|@{}c@{}|@{}c@{}|@{}c@{}|@{}c@{}|}
\hline	\textbf{\#}	&	\textbf{van Gogh}	&	\textbf{Rembrandt}	&	\textbf{Renoir}	\\
\hline
\hline	1	&	bandaged-ear	&	abraham	&	apres-bain \\
\hline	2	&	berceuse	&	anslo	&	baigneuses \\
\hline	3	&	cordeville	&	aristotle-homer	&	bathers-1887 \\
\hline	4	&	corridor-asylum	&	artemis	&	bathers-1918 \\
\hline	5	&	cypress-star	&	bathing-river	&	bougival \\
\hline	6	&	cypresses	&	bathsheba	&	canoeist \\
\hline	7	&	flower-beds-holland	&	belshazzar	&	chocquet \\
\hline	8	&	green-vineyard	&	children	&	city \\
\hline	9	&	green-wheat-field	&	danae	&	country \\
\hline	10	&	house-ploughman	&	david	&	dancer \\
\hline	11	&	mme-trabuc	&	descent	&	durieux \\
\hline	12	&	mr-trabuc	&	emmaus	&	flowers \\
\hline	13	&	old-mill	&	hendrickje	&	gabrielle \\
\hline	14	&	old-vineyard	&	holy-family	&	girl-seated \\
\hline	15	&	olive-alpilles	&	jan-six	&	jugglers \\
\hline	16	&	olive-trees	&	magn-glass	&	lady-piano \\
\hline	17	&	orchard-bloom-poplars	&	meditation	&	laundress \\
\hline	18	&	orchard-plum-trees	&	mill	&	loge \\
\hline	19	&	poppies	&	music-party	&	lucie-berard \\
\hline	20	&	red-vineyard	&	nicolaes-tulp	&	near-lake \\
\hline	21	&	reminiscences	&	old-man	&	fournaise \\
\hline	22	&	road-menders	&	ostrich	&	horsewoman \\
\hline	23	&	roulin	&	potiphar	&	meadow \\
\hline	24	&	self-1	&	prodigal-son	&	moulin-galette \\
\hline	25	&	self-2	&	raising-lazarus	&	nini \\
\hline	26	&	self-easel	&	.1640	&	parapluies \\
\hline	27	&	self-gauguin	&	.1661	&	premiere-sortie \\
\hline	28	&	self-orsay	&	.1669	&	promenade \\
\hline	29	&	self-whitney	&	.night-watch	&	ride \\
\hline	30	&	skull-cigarette	&	return-prodigal-son	&	romain-lacaux \\
\hline	31	&	sun-cloud	&	ruts	&	sisley-wife \\
\hline	32	&	threatening-skies	&	samson	&	women \\
\hline	33	&	trees-asylum	&	scholar	&	seashore \\
\hline	34	&	trees-ivy-asylum	&	self-1629	&	seated-bather \\
\hline	35	&	village-stairs	&	self-1634	&	sewing \\
\hline	36	&	wheat-field	&	self-1660	&	sisley \\
\hline	37	&	wheat-rising-sun	&	slaughtered-ox	&	swing \\
\hline	38	&	willows	&	staalmeesters	&	terrace \\
\hline	39	&	peasant	&	stofells	&	watercan \\
\hline	40	&	woman-arles	&	tobias	&	woman-veil \\

\hline 
\end{tabular} 
\end{document}